%% file: main.tex
\documentclass[11pt]{article}

\usepackage[final]{acl}

\usepackage{times}
\usepackage{latexsym}

\usepackage[T1]{fontenc}
\usepackage[utf8]{inputenc}

\usepackage{microtype}

\usepackage{inconsolata}

\usepackage{graphicx}
\usepackage{amsmath}
\usepackage{booktabs}

\usepackage{pdfpages}
\usepackage{algorithm}
\usepackage{algorithmic}
\usepackage{longtable}

\title{DementiaBank-Emotion: A Multi-Rater Emotion Annotation Corpus for Alzheimer's Disease Speech (Version 1.0)}

\author{
  \textbf{Cheonkam Jeong\textsuperscript{1,7,*}},
  \textbf{Jessica Liao\textsuperscript{2}},
  \textbf{Audrey Lu\textsuperscript{2}},
  \textbf{Yutong Song\textsuperscript{2}},
  \textbf{Christopher Rashidian\textsuperscript{2}},
  \\
  \textbf{Donna Krogh\textsuperscript{3}},
  \textbf{Erik Krogh\textsuperscript{3}},
  \textbf{Mahkameh Rasouli\textsuperscript{1}},
  \textbf{Jung-Ah Lee\textsuperscript{1}},
  \textbf{Nikil Dutt\textsuperscript{2}},
  \\
  \textbf{Lisa Gibbs\textsuperscript{6}},
  \textbf{David Sultzer\textsuperscript{4}},
  \textbf{Julie Rousseau\textsuperscript{6}},
  \textbf{Jocelyn Ludlow\textsuperscript{1}},
  \textbf{Margaret Galvez\textsuperscript{2}},
  \\
  \textbf{Alexander Nuth\textsuperscript{8}},
  \textbf{Chet Khay\textsuperscript{5}},
  \textbf{Sabine Brunswicker\textsuperscript{7,$\dagger$}},
  \textbf{Adeline Nyamathi\textsuperscript{1,$\dagger$}}
  \\
  \\
  \small{\textsuperscript{1}Sue \& Bill Gross School of Nursing, University of California, Irvine (UCI)} \\
  \small{\textsuperscript{2}Donald Bren School of Information and Computer Sciences, UCI, USA} \\
  \small{\textsuperscript{3}Smart Forward, Rancho Palos Verdes, USA} \\
  \small{\textsuperscript{4}Dept. of Psychiatry and Human Behavior, UCI, USA} \\
  \small{\textsuperscript{5}Amore Senior Living, Laguna Niguel, USA} \\
  \small{\textsuperscript{6}School of Medicine, University of California, Irvine} \\
  \small{\textsuperscript{7}Purdue University, West Lafayette, USA} \\
  \small{\textsuperscript{8}The George Washington University, Washington, D.C., USA}
}

\begin{document}
\maketitle

\begin{abstract}
We present DementiaBank-Emotion, the first multi-rater emotion annotation corpus for Alzheimer's disease (AD) speech. Annotating 1,492 utterances from 108 speakers for Ekman's six basic emotions and \textit{neutral}, we find that AD patients express significantly \textit{more} non-neutral emotions (16.9\%) than healthy controls (5.7\%; $p$ < .001). Exploratory acoustic analysis suggests a possible dissociation: control speakers showed substantial F0 modulation for sadness ($\Delta$ = $-$3.45 semitones from baseline), whereas AD speakers showed minimal change ($\Delta$ = +0.11 semitones; interaction $p$ = .023), though this finding is based on limited samples (sadness: $n$=5 control, $n$=15 AD) and requires replication. Within AD speech, loudness differentiates emotion categories, indicating partially preserved emotion-prosody mappings. We release the corpus, annotation guidelines, and calibration workshop materials to support research on emotion recognition in clinical populations.
\end{abstract}

\section{Introduction}
Dementia is characterized not only by cognitive decline but also by progressive impairments in language and communication \citep{mueller2018connected}. Computational approaches to AD speech analysis have made significant progress, with studies identifying linguistic markers such as reduced lexical diversity, simplified syntax, and increased pauses \citep{fraser2016linguistic, ahmed2013connected}. The ADReSS Challenge \citep{luz2020adress} established benchmarks for automatic AD detection using acoustic and linguistic features from the DementiaBank Pitt Corpus \citep{becker1994natural}.

However, the \textit{affective} dimension of AD speech remains underexplored. Emotion recognition in speech has received considerable attention in the NLP and speech processing communities, with datasets such as IEMOCAP \citep{busso2008iemocap} enabling research on multimodal emotion recognition in conversations. Yet these resources focus on healthy populations, typically using acted or scripted speech. No comparable resource exists for clinical populations with cognitive impairment.

Understanding emotional expression in AD speech is important for several reasons. First, emotion may serve as an additional marker for disease progression, complementing existing linguistic and acoustic features \citep{henry2009emotion}. Second, caregivers and clinicians need to accurately interpret emotional cues from AD patients, which may differ from typical patterns. Third, emotion-aware assistive technologies require training data that reflects the actual emotional expressions of target users.

In this paper, we present \textbf{DementiaBank-Emotion} (v1.0), a multi-rater emotion annotation corpus for AD speech. Our contributions are:

\begin{enumerate}
    \item We release the first emotion-annotated corpus for AD speech: 1,492 utterances from 108 speakers with multi-rater labels (Fleiss' $\kappa$ = 0.23--0.31 post-calibration).
    \item We document annotation challenges specific to clinical speech and provide calibration workshop materials addressing issues such as distinguishing face-saving laughter from joy.
    \item We find that AD patients express \textit{more} non-neutral emotions than controls (16.9\% vs.\ 5.7\%), with exploratory evidence suggesting reduced prosodic differentiation (acoustic flattening) that warrants further investigation.
    \item We show that loudness differentiates emotion categories within AD speech, suggesting partially preserved emotion-prosody mappings.
\end{enumerate}

\section{Related Work}

\subsection{Emotion Recognition in Spontaneous Speech}

Benchmark datasets such as IEMOCAP \citep{busso2008iemocap}, MSP-IMPROV \citep{busso2017mspimprov}, and RAVDESS \citep{livingstone2018ravdess} have provided foundational resources for speech emotion recognition (SER). However, these corpora predominantly rely on acted or semi-scripted speech from healthy individuals, which often feature prototypical and exaggerated emotional expressions.

Spontaneous clinical speech, particularly from patients with Alzheimer's Disease (AD), poses unique challenges that these datasets do not address. In AD, the mapping between acoustic markers and emotional states is often confounded by pathological changes in voice quality, such as flat affect or reduced $F_0$ variance \citep{kreiman1993perceptual}. Furthermore, linguistic focus and pragmatic intent and function\footnote{In this context, ``pragmatic'' is used as a general descriptive term rather than a strictly bound technical term within a specific linguistic framework.}, may be critical for identifying emotions like surprise, are frequently disrupted in cognitive decline, requiring a more nuanced annotation framework than standard categorical labeling.

\subsection{Acoustic and Linguistic Analysis of AD Speech}

The DementiaBank Pitt Corpus \citep{becker1994natural, lanzi2023dementiabank} has been extensively utilized for AD classification, primarily focusing on idea density, lexical diversity, and disfluency patterns \citep{fraser2016linguistic, luz2020adress}. While voice quality features and the eGeMAPS set \citep{eyben2016geneva} have successfully distinguished AD and MCI from healthy controls \citep{themistocleous2020voice}, these studies typically treat acoustic features as diagnostic markers for cognitive status rather than as vehicles for emotional expression.

A critical gap exists in understanding the \textit{affective} dimensions of AD speech. Standard sentiment analysis and automated Speech Emotion Recognition (SER) systems often fail to account for clinical nuances—such as laughter serving as a coping mechanism for word-finding difficulties rather than as a signal of joy \citep{glenn2003laughter}. Recent work by \citet{chou2025multimodal} explored multimodal AD classification combining facial and eye-tracking data with emotion prediction, finding that emotion features alone did not significantly improve AD detection accuracy. This suggests that the relationship between emotion and cognitive status may be more nuanced than simple feature concatenation can capture. Our work addresses this gap by introducing a multi-rater annotation layer that explicitly incorporates linguistic focus, pragmatic intent, and clinical expertise, enabling fine-grained analysis of \textit{how} emotions are expressed rather than merely \textit{whether} they are present.

\section{Corpus Description}

\subsection{Source Data}

The DementiaBank Pitt Corpus \citep{becker1994natural, lanzi2023dementiabank} contains audio recordings and transcripts of English-speaking participants completing the Cookie Theft picture description task from the Boston Diagnostic Aphasia Examination. In this task, participants describe a scene depicting a kitchen where a woman washes dishes while water overflows from the sink, and children steal cookies from a jar while one child's stool tips over.

For this corpus (v1.0), both AD patient and control data were drawn from the ADReSS 2020 Challenge training set \citep{luz2020adress}, which provides a cross-sectional subset of the longitudinal DementiaBank data with matched AD and control groups (54 speakers each, balanced for age and gender). A future version (v2.0) will include the full DementiaBank longitudinal data.

\subsection{Corpus Statistics}

Table~\ref{tab:corpus_stats} summarizes the corpus. We annotated utterances from 108 speakers: 54 AD patients and 54 healthy controls from the ADReSS Challenge dataset, which was designed with matched demographics (age, gender) between groups. For emotion analysis, we focus on participant (PAR) utterances; investigator (INV) utterances are predominantly neutral prompts and backchannels.

Of the 1,492 total PAR utterances, 615 AD and 731 control utterances received valid final emotion labels. The remaining 137 AD and 9 control utterances were marked as \textit{ambiguous} due to annotator disagreement that could not be resolved through the adjudication algorithm; these were excluded from emotion distribution analysis but retained for acoustic analysis where applicable.

\begin{table}[t]
\centering
\small
\begin{tabular}{lrr}
\toprule
 & \textbf{AD} & \textbf{Control} \\
\midrule
Speakers & 54 & 54 \\
Sessions & 54 & 54 \\
PAR utterances & 752 & 740 \\
\quad Valid labels & 615 & 731 \\
\quad Ambiguous & 137 & 9 \\
\bottomrule
\end{tabular}
\caption{Corpus statistics by group.}
\label{tab:corpus_stats}
\end{table}

\section{Annotation}
\subsection{Annotators}
The annotation involved 11 raters with diverse expertise. The first round was conducted by clinical experts (four Ph.D.-level nursing researchers and one nurse practitioner). The second round included three clinical experts and a professor of business. The third round and control data annotation were performed by a technical team, including computer science and polytechnic researchers, under the supervision of the lead clinical expert. This multidisciplinary composition ensured that the labels reflect both clinical reality and structural consistency for computational analysis.

\subsection{Annotation Scheme}
We adopted Ekman's basic emotion categories \citep{ekman1992basic} plus neutral, yielding seven labels: neutral, joy, sadness, fear, anger, surprise, and disgust. To ensure conceptual consistency, our annotation guidelines incorporated the formal definitions provided by the Paul Ekman Group\footnote{\url{https://www.paulekman.com/universal-emotions/}}. Annotators were trained to recognize not only the prototypical vocal expressions but also the underlying psychological themes associated with each category—for instance, identifying \textit{sadness} through themes of loss or helplessness, and \textit{surprise} as a reaction to unexpected visual stimuli in the Cookie Theft picture. This dual focus on both acoustic properties and the theoretical definitions of universal emotions provided a robust framework for interpreting the complex affective signals in AD speech.

\begin{table}[t]
\centering
\small
\begin{tabular}{lcccc}
\toprule
\textbf{Emotion} & \textbf{F0} & \textbf{F0 var.} & \textbf{Loudness} & \textbf{Rate} \\
\midrule
Joy & -- & High & High & Moderate \\
Sadness & Low & Low & Low & Slow \\
Fear & High & High & High & Fast \\
Anger & Low & Low & High & Fast \\
Disgust & Low & Low & Low & Slow \\
Surprise & \multicolumn{4}{c}{Dramatic pitch change, gasp} \\
Neutral & \multicolumn{4}{c}{Flat, monotonous} \\
\bottomrule
\end{tabular}
\caption{Expected prosodic cues by emotion category, based on annotation guidelines adapted from \citet{sobin1999emotion}.}
\label{tab:prosody_cues}
\end{table}

\subsection{Annotation Guidelines}
The initial annotation framework was developed by the lead researcher (Ph.D. in Linguistics specializing in Phonology and Phonetics). This framework integrated Ekman's psychological emotion categories with objective acoustic markers, such as $F_0$ variance, volume thresholds, and phonation types \citep{ekman1992basic, sobin1999emotion}. Table~\ref{tab:prosody_cues} summarizes the expected prosodic cues for each emotion category.

\subsection{Calibration Workshops}
\label{sec:workshop}
To address discrepancies observed during the initial labeling phase, we conducted multiple calibration workshops. These sessions included the annotators and an advisory panel consisting of a Professor of Psychiatry and the Director of a memory care facility. The workshops functioned as an iterative feedback loop: after each batch of annotation, the lead researcher provided feedback and the panel discussed ambiguous cases. 

A primary outcome of these discussions was the refinement of the \textbf{Default to Neutral} principle, specifically considering the task's nature as a picture description. We observed that annotators were often influenced by charged lexical items (e.g., ``robbing'' a cookie jar) even when the speaker's delivery was flat. The panel reached a consensus to prioritize prosodic cues over lexical content, leading to the development of the formalized Guideline v2.0 (see Appendix~\ref{app:appendix_guideline}) formed their final rounds of labeling independently.

\begin{table*}[t]
\centering
\footnotesize
\begin{tabular}{@{}p{4cm}llccccccc@{}}
\toprule
& & & & \multicolumn{2}{c}{\textbf{F0 (st)}} & \multicolumn{2}{c}{\textbf{F0 var}} & \multicolumn{2}{c}{\textbf{Loudness}} \\
\cmidrule(lr){5-6} \cmidrule(lr){7-8} \cmidrule(lr){9-10}
\textbf{Utterance} & \textbf{Label} & \textbf{Cues} & \textbf{Spk} & eGe & spk & eGe & $z$ & eGe & $z$ \\
\midrule
``and his stool is about to dump him \&=laughs'' & Joy & happy laugh & 234 (M, 66) & 24.2 & $-$2.8 & 0.26 & +0.27 & 0.66 & +0.50 \\
``well the poor mother's a-doin(g) dishes \&=laughs'' & Sad & helpless laugh & 337 (F, 59) & 33.2 & +0.4 & 0.15 & $-$0.69 & 0.38 & $-$0.62 \\
``he's about to drop off that stool too'' & Surp & stress on ``stool'' & 234 (M, 66) & 25.8 & $-$1.2 & 0.28 & +0.78 & 0.41 & $-$0.57 \\
``well the kids is robbin(g) a cookie jar'' & Neut & flat tone, factual & 234 (M, 66) & 25.8 & $-$1.2 & 0.18 & $-$1.28 & 0.93 & +1.71 \\
\bottomrule
\end{tabular}
\caption{Calibration workshop examples with agreed labels, diagnostic cues, and acoustic features. Spk = Speaker ID (sex, age). eGe = eGeMAPS values (F0 in
semitones from 27.5 Hz; F0 var = stddevNorm; Loudness in sone). spk/$z$ = speaker-normalized (F0 in semitones from speaker mean; others $z$-scored within
speaker). Acoustic values were calculated post-annotation.}
\label{tab:workshop}
\end{table*}

\paragraph{Laughter Subtypes.} Not all laughter indicates joy. Workshop discussions distinguished (1) \textit{happy laughter} accompanying genuine amusement (labeled as joy), (2) \textit{helpless laughter} co-occurring with expressions of sympathy such as ``poor'' (labeled as sadness), and (3) \textit{sarcastic laughter} (context-dependent). This distinction proved critical for clinical speech, where laughter often serves face-saving or coping functions.

\paragraph{Prosodic Focus and Surprise.} Following prosodic theories of focus \citep{bolinger1958stress, selkirk1995sentence}, utterances with dramatic pitch changes ($F_0$ excursions), gasps, or strong stress on unexpected elements (e.g., ``he's about to drop off that \textsc{stool}'') were labeled as \textit{surprise}, as illustrated in Table~\ref{tab:workshop}. In these instances, the acoustic prominence functions as a marker of pragmatic focus and new information (information focus) ($z$ = +0.78). This indicates that the speaker has cognitively shifted from routine description to actively ``noticing'' a salient, unanticipated element in the scene.

\paragraph{Default to Neutral.} Flat intonation with factual, descriptive content was the primary cue for neutrality. Even when lexical items might suggest emotion (e.g., ``robbing''), annotators labeled utterances as neutral if prosodic delivery was flat.

\subsection{Inter-Rater Reliability}

Table~\ref{tab:irr} shows inter-rater reliability across annotation batches. For patient data, Fleiss' $\kappa$ improved from 0.094 (Batch 1, before calibration workshop) to 0.313 (Batch 2) after the workshop established consensus on ambiguous cases.

\begin{table}[h]
\centering
\small
\begin{tabular}{lcc}
\toprule
\textbf{Data} & \textbf{Fleiss' $\kappa$} & \textbf{$n$ raters} \\
\midrule
Patient Batch 1 & 0.094 & 5 \\
Patient Batch 2 & 0.313 & 5 \\
Patient Batch 3 & 0.231 & 5 \\
Control (1 excl.) & 0.254 & 3 \\
\bottomrule
\end{tabular}
\caption{Inter-rater reliability by annotation batch. For control data, one rater was excluded due to divergent labeling patterns (see text).}
\label{tab:irr}
\end{table}

For control data, annotation proceeded without a separate calibration workshop due to time constraints; instead, we collected written feedback from annotators. Two of the four annotators had prior experience from the patient annotation task, while two annotators (L3 and L4) were newly recruited. Non-neutral labeling rates varied substantially across annotators (ranging from 5\% to 26\%); excluding both new annotators would have left insufficient raters, so we excluded only L3 based on pairwise agreement analysis, yielding $\kappa$ = 0.254 for the remaining 3 raters.

Written feedback revealed a critical difference in annotation strategies. Experienced annotators explicitly reported attending to \textit{within-speaker variation}: L1 noted ``I labeled based on change of their typical pattern,'' and L2 reported ``I focused on differences in speech pitch/speed/tone between utterances.'' In contrast, a newly recruited annotator reported relying heavily on explicit guidelines (``90\% guideline-based, only 10\% perception''), suggesting an imbalance between rule-following and perceptual judgment.

This divergence illustrates that effective emotion annotation requires integrating explicit criteria with implicit perceptual skills---particularly the ability to detect within-speaker deviations from baseline affect. Calibration workshops appear to foster this integration not merely by clarifying guidelines, but by developing shared intuitions through discussion of ambiguous cases. Without such calibration, annotators may default to either extreme: over-reliance on guidelines (missing subtle within-speaker variations) or over-reliance on perception (applying idiosyncratic criteria). Future annotation efforts should include calibration sessions for all data to establish this balance.

While these reliability values are moderate by conventional standards, they are comparable to IEMOCAP ($\kappa$ = 0.27 for categorical labels; \citealp{busso2008iemocap}) and reflect the inherent difficulty of emotion annotation in clinical speech \citep{kreiman1993perceptual}. We note that control data had only four annotators (three after exclusion), which may affect comparability with the five-rater patient annotation.

\subsection{Final Label Determination}

\begin{algorithm}[h]
\small
\caption{Hierarchical Label Determination}
\label{alg:label_determination}
\begin{algorithmic}[1]
\REQUIRE $L$: list of labels from $n$ annotators, $C$: confidence scores
\ENSURE $GoldLabel$: final assigned label

\IF{majority count of any label $x \in L \geq \lceil n/2 \rceil$}
    \STATE $GoldLabel \gets x$
\ELSIF{tie exists between \textit{Neutral} and \textit{Emotion} $e$}
    \STATE $GoldLabel \gets e$ \COMMENT{Non-neutral preference}
\ELSE
    \STATE $W_i \gets$ calculate weighted sum for each tied label $i$ using $C$
    \IF{max($W$) is unique}
        \STATE $GoldLabel \gets \text{argmax}(W)$
    \ELSE
        \STATE $GoldLabel \gets \text{NaN}$ \COMMENT{Marked as ambiguous}
    \ENDIF
\ENDIF
\end{algorithmic}
\end{algorithm}

To determine the gold standard label for each utterance, we implemented a hierarchical adjudication process as described in Algorithm~\ref{alg:label_determination}. This procedure prioritizes consensus and emotional sensitivity while utilizing confidence scores to resolve ties.

For AD patient data, all five annotators' labels were used with a majority threshold of $\geq$3. For control data, L3's annotations were excluded due to the divergent labeling pattern described above, and the same algorithm was applied to the remaining three annotators with a majority threshold of $\geq$2.

Utterances resulting in \textit{NaN} (137 in AD, 9 in control) were classified as ambiguous and analyzed separately in Section~\ref{sec:ambiguous}.

\section{Corpus Analysis}

\subsection{Emotion Distribution}
\begin{figure}[t]
\centering
\includegraphics[width=\columnwidth]{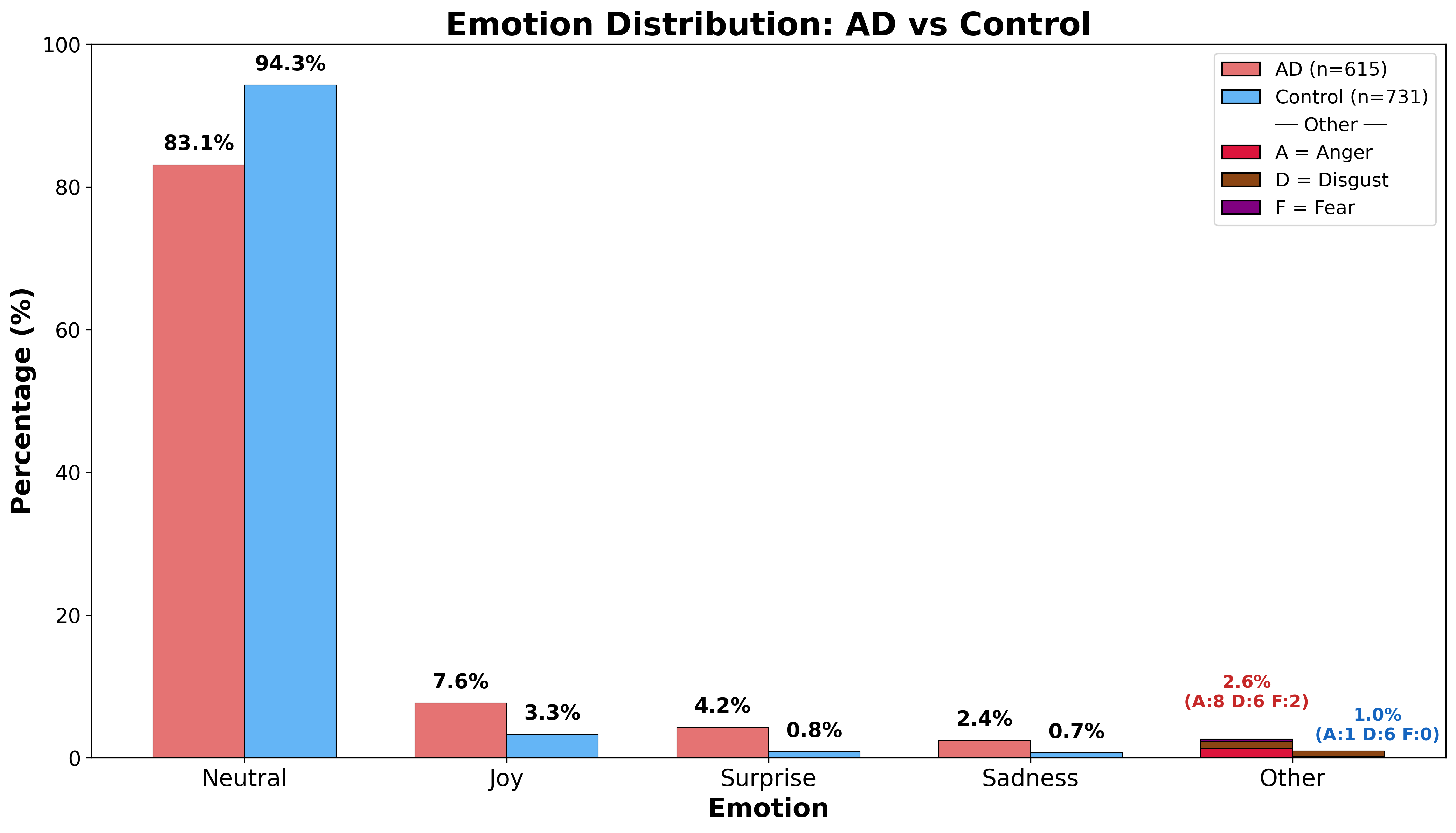}
\caption{Emotion distribution comparing AD patients ($n$=615) and healthy controls ($n$=731). Rare emotions (anger, disgust, fear) were pooled as ``Other'' with counts shown. AD patients show significantly higher rates of non-neutral emotions (16.9\% vs.\ 5.7\%; $\chi^2$ = 38.45, $p < .001$).}
\label{fig:emotion_dist}
\end{figure}

Table~\ref{tab:emotion_dist} and Figure~\ref{fig:emotion_dist} show the emotion distribution for labeled PAR utterances. AD patients exhibited significantly higher rates of non-neutral emotions (16.9\%) compared to controls (5.7\%; $\chi^2$(1) = 38.45, $p < .001$).

\textit{Joy} was the most frequent non-neutral emotion in both groups (AD: 7.6\%, Control: 3.3\%), followed by \textit{surprise} (AD: 4.2\%, Control: 0.8\%). Sadness, anger, disgust, and fear were rare overall but more frequent in AD speech.

\begin{table}[t]
\centering
\small
\begin{tabular}{lrrrr}
\toprule
 & \multicolumn{2}{c}{\textbf{AD}} & \multicolumn{2}{c}{\textbf{Control}} \\
\textbf{Emotion} & $n$ & \% & $n$ & \% \\
\midrule
Neutral & 511 & 83.1 & 689 & 94.3 \\
Joy & 47 & 7.6 & 24 & 3.3 \\
Surprise & 26 & 4.2 & 6 & 0.8 \\
Sadness & 15 & 2.4 & 5 & 0.7 \\
Anger & 8 & 1.3 & 1 & 0.1 \\
Disgust & 6 & 1.0 & 6 & 0.8 \\
Fear & 2 & 0.3 & 0 & 0.0 \\
\midrule
\textbf{Non-neutral} & 104 & 16.9 & 42 & 5.7 \\
\midrule
\textbf{Total} & 615 & 100 & 731 & 100 \\
\bottomrule
\end{tabular}
\caption{Emotion distribution for labeled PAR utterances.}
\label{tab:emotion_dist}
\end{table}

\subsection{Ambiguous Cases}
\label{sec:ambiguous}

Of the 146 utterances marked as ambiguous, 137 occurred in AD speech and 9 in control speech. Comparing ambiguous versus non-ambiguous AD utterances revealed that ambiguous cases had significantly lower F0, F0 variance, loudness, and HNR, as well as fewer words---suggesting that ambiguity often arises from insufficient prosodic cues.

Analysis revealed two patterns of ambiguity: (1) the majority exhibited flat or reduced acoustic profiles, providing insufficient prosodic cues for emotion identification; (2) a subset showed elevated F0 variance compatible with multiple emotion categories. Table~\ref{tab:ambiguous} illustrates the latter pattern.

\begin{table*}[h]
\centering
\footnotesize
\begin{tabular}{@{}p{6cm}ccccccc@{}}
\toprule
& & \multicolumn{2}{c}{\textbf{F0 (st)}} & \multicolumn{2}{c}{\textbf{F0 var}} & \multicolumn{2}{c}{\textbf{Loudness}} \\
\cmidrule(lr){3-4} \cmidrule(lr){5-6} \cmidrule(lr){7-8}
\textbf{Utterance} & \textbf{Spk} & eGe & spk & eGe & $z$ & eGe & $z$ \\
\midrule
``she don't know what the hell to think of it'' & 010 (M, 69) & 21.0 & $-$0.1 & 0.28 & +1.04 & 0.17 & $-$0.47 \\
``one of the kids is gonna get a crack on'' & 018 (M, 66) & 31.4 & $-$4.2 & 0.29 & +0.95 & 0.07 & $-$0.58 \\
\bottomrule
\multicolumn{8}{l}{\scriptsize Labels: (top) disg(2), ang(2), sad(1); (bottom) fear, surp, neut, joy} \\
\end{tabular}
\caption{Representative ambiguous utterances (elevated F0 variance type). eGe = eGeMAPS values; spk/$z$ = speaker-normalized.}
\label{tab:ambiguous}
\end{table*}

These findings suggest that ambiguity in AD emotion annotation arises from two sources: insufficient prosodic information due to flat affect, and prosodic cues that are genuinely compatible with multiple emotional interpretations.

\subsection{Qualitative Analysis}
Examination of non-neutral utterances reveals discourse patterns consistent with the annotation criteria established in our calibration workshops (Section~\ref{sec:workshop}).

\paragraph{Laughter as Face-Saving.} Many ``joy'' labels (47 AD utterances) co-occurred with laughter tokens (e.g., ``\&=laughs''). As discussed in Section~\ref{sec:workshop}, laughter subtypes were distinguished by prosodic cues: happy laughter with rising intonation was labeled joy, while helpless laughter with falling tone was labeled sadness. This distinction proved critical, as laughter in AD speech often serves a face-saving function during moments of communicative difficulty rather than expressing genuine happiness.

\paragraph{Surprise and Noticing.} Surprise labels (26 AD utterances) frequently occurred when participants noticed chaotic elements of the picture, often marked by the interjection ``oh'' and accompanied by pitch excursions.

\paragraph{Sadness and Empathy.} Sadness labels (15 AD utterances) were characterized by falling intonation and slower tempo, often co-occurring with expressions of sympathy toward depicted characters. These utterances suggest preserved emotional engagement with narrative content.

\paragraph{Anger and Frustration.} Anger labels (8 AD utterances) reflected task-related frustration, as in ``what the hell else?'' and ``I don't know!'' Such expressions may indicate awareness of communicative difficulty.

\paragraph{Clinical Implications.} These patterns imply that emotion annotation in AD speech requires attention to pragmatic and interactional functions, not merely surface cues. Laughter-as-coping, frustration-as-self-awareness, and empathy-toward-characters all reflect preserved social-emotional processing that may coexist with linguistic impairment.

\subsection{Linguistic and Acoustic Analysis}

Table~\ref{tab:features} shows speaker-level linguistic and acoustic features. AD patients produced marginally shorter utterances (MLU = 7.84 words) compared to controls (MLU = 8.70; $t$ = $-$2.13, $p$ = .035, $d$ = $-$0.41). Total words per session and type-token ratio (TTR) did not differ significantly between groups.

Acoustic features were extracted using openSMILE \citep{eyben2010opensmile} with the eGeMAPS feature set \citep{eyben2016geneva}, which includes prosodic features (F0, loudness), voice quality measures (jitter, shimmer, HNR), and spectral parameters. Notably, H1-H2 (the amplitude difference between the first and second harmonics) indexes phonation type: higher values indicate breathier voice quality, while lower values suggest pressed or creaky phonation \citep{gordon2001phonation,hanson1997glottal,keating2015acoustic}.

AD patients showed significantly higher HNR ($d$ = 0.42, $p$ = .033) compared to controls, though this difference may reflect variations in recording conditions rather than voice quality. Other acoustic features (F0, loudness, shimmer, H1-H2) did not differ significantly between groups at the speaker level. The lack of group differences in most acoustic features contrasts with prior work using smaller control samples; our matched cohort (54 speakers per group) provides a more balanced comparison.

\begin{table*}[t]
\centering
\small
\setlength{\tabcolsep}{4pt}
\begin{tabular}{@{}lcccccccc@{}}
\toprule
& \multicolumn{3}{c}{\textbf{Linguistic ($n$=54 AD, 54 Ctrl)}} & \multicolumn{5}{c}{\textbf{Acoustic ($n$=54 AD, 54 Ctrl)}} \\
\cmidrule(lr){2-4} \cmidrule(lr){5-9}
\textbf{Group} & MLU & Words & TTR & Loud. & Shim. & HNR & F0 & H1-H2 \\
\midrule
AD & 7.84 (2.17) & 106.9 (67.4) & 0.58 (0.10) & 0.41 (0.37) & 1.40 (0.26) & 4.89 (2.29) & 28.63 (3.89) & 2.69 (3.90) \\
Control & 8.70 (2.06) & 118.3 (51.8) & 0.59 (0.07) & 0.50 (0.45) & 1.45 (0.28) & 3.96 (2.19) & 27.16 (4.99) & 3.39 (3.46) \\
\midrule
$p$ & .035* & .328 & .422 & .291 & .296 & .033* & .091 & .326 \\
$d$ & $-$.41 & $-$.19 & $-$.16 & $-$.20 & $-$.20 & .42 & .33 & $-$.19 \\
\bottomrule
\end{tabular}
\caption{Linguistic and acoustic features (speaker-level). Values are Mean (SD). Shimmer (Shim.) and HNR in dB, F0 in semitones, Loudness (Loud.) in sones. *$p<.05$, **$p<.01$, ***$p<.001$.}
\label{tab:features}
\end{table*}

\subsection{Emotion and Acoustics}

\paragraph{F0 Normalization.} F0 was converted to semitones using each speaker's mean F0 as the reference \citep{JEONG2026101474,rose2002forensic,nolan2009variation}:
\begin{equation}
\text{F0}_{\text{st}} = 12 \times \log_2\left(\frac{\text{F0}_{\text{Hz}}}{\text{F0}_{\text{mean,speaker}}}\right)
\end{equation}
This speaker-relative normalization isolates intonational dynamics from physiological baselines, enabling comparison across speakers. Other acoustic features were z-scored within speaker.

\begin{figure}[t!]
\centering
\includegraphics[width=\columnwidth]{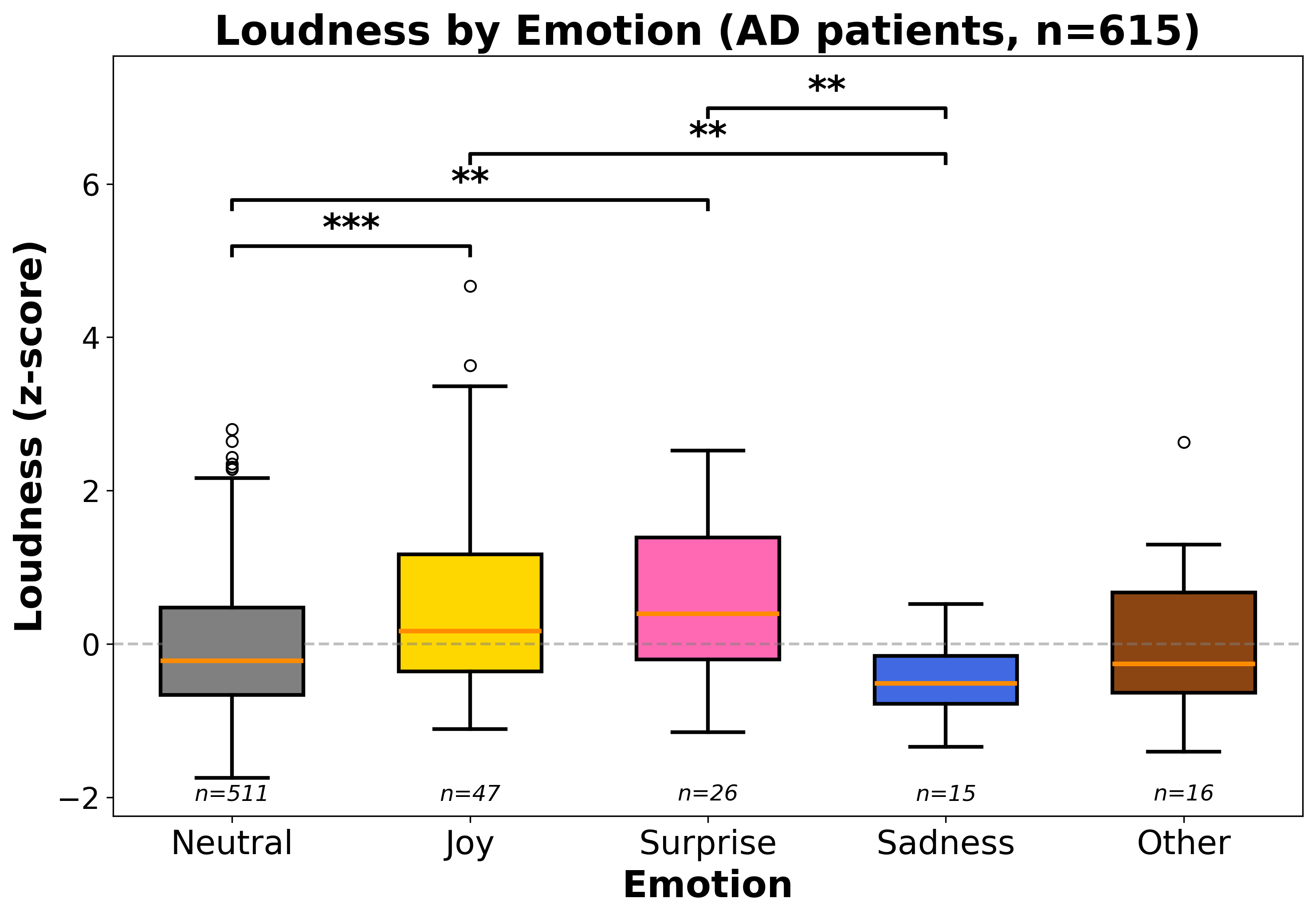} 

\caption{Speaker-normalized loudness by emotion category in AD patients. Rare emotions (anger, disgust, fear) were pooled as ``Other.'' Asterisks indicate significant pairwise differences (Tukey HSD): **$p<.01$, ***$p<.001$. Other acoustic features (F0, jitter, shimmer, HNR) showed no significant differences (see Appendix~\ref{app:acoustic_emotion}).}
\label{fig:acoustic_emotion}
\end{figure}

Within AD patients, we examined whether acoustic features differentiate emotion categories using these speaker-normalized values. Rare emotions (anger, disgust, fear; each $n<10$) were pooled into an ``Other'' category ($n$=16) for statistical power. One-way ANOVA revealed a significant effect of emotion on loudness ($F$(4, 610) = 8.48, $p$ < .001, $\eta^2$ = 0.053), but not on F0 ($F$(4, 610) = 2.11, $p$ = .078). Post-hoc Tukey HSD tests showed that joy and surprise utterances had significantly higher loudness than neutral and sadness: joy vs.\ neutral ($\Delta$ = +0.59, $p$ < .001), surprise vs.\ neutral ($\Delta$ = +0.64, $p$ = .004), joy vs.\ sadness ($\Delta$ = +1.00, $p$ = .002), and surprise vs.\ sadness ($\Delta$ = +1.04, $p$ = .003). Figure~\ref{fig:acoustic_emotion} visualizes these patterns. Other acoustic features (F0, jitter, shimmer, HNR) did not show significant differences by emotion after speaker normalization.

\paragraph{Group $\times$ Emotion Interaction.} To examine whether AD and control speakers differ in how they acoustically realize emotions, we fitted linear mixed-effects models with speaker as a random intercept: Feature $\sim$ Group $\times$ Emotion + (1$|$Speaker). A significant Group $\times$ Sadness interaction emerged for F0 ($\beta$ = $-$5.20, SE = 2.28, $p$ = .023): when expressing sadness, control speakers showed a substantial F0 decrease relative to their neutral baseline ($\Delta$ = $-$3.45 semitones), whereas AD speakers showed virtually no F0 modulation ($\Delta$ = +0.11 semitones). Bootstrap resampling at the speaker level (1,000 iterations) yielded a 95\% confidence interval of [$-$5.70, $-$0.22] for the interaction effect, confirming statistical significance. A similar pattern was observed for shimmer ($\beta$ = $-$0.53, $p$ = .017). These findings suggest that while AD patients express more emotional content (16.9\% vs.\ 5.7\% non-neutral), their prosodic realization of these emotions is attenuated---a pattern we term \textit{acoustic flattening}.

\section{Discussion}

\subsection{Interpreting Higher Emotional Expression in AD}

Our finding that AD patients express significantly more non-neutral emotions (16.9\%) than healthy controls (5.7\%; $\chi^2$ = 38.45, $p < .001$) warrants careful interpretation. At first glance, this pattern might seem counterintuitive: one might expect cognitive decline to manifest as emotional flattening or reduced expressivity. However, our analysis suggests that emotional expression in AD speech reflects preserved social-emotional processing that operates alongside---and sometimes compensates for---linguistic impairment.

As documented in Section~\ref{sec:workshop}, laughter in AD speech often serves a face-saving function rather than expressing genuine happiness. This finding has important implications for emotion recognition systems: a naive classifier trained on healthy speech might map laughter tokens directly to joy, systematically misclassifying these compensatory instances. The higher rate of non-neutral expression in AD speech may therefore reflect the increased deployment of emotional expressions as coping strategies during moments of linguistic difficulty \citep{glenn2003laughter}.

\subsection{The Cookie Theft Task as Emotional Elicitor}

The Cookie Theft picture description task, while designed primarily as a cognitive-linguistic assessment, creates conditions that may naturally elicit emotional responses---particularly for individuals experiencing word-finding difficulties. The task places speakers in a public performance context where they must demonstrate competence, and the chaotic scene depicted provides ample material for emotional engagement.

Our analysis revealed that AD patients remain emotionally responsive to their environment: surprise marked moments of noticing unexpected elements, while sadness reflected empathy toward depicted characters. These capacities may be clinically significant for understanding quality of life and social functioning in AD.

\subsection{Acoustic Signatures of Emotion in AD Speech}

Within AD patients, loudness significantly differentiated emotion categories, with joy and surprise utterances showing higher loudness than neutral. These patterns are broadly consistent with the acoustic correlates of emotion documented in healthy populations, suggesting that the mapping between emotion and acoustic expression is at least partially preserved in AD.

An exploratory analysis comparing AD and control groups revealed a suggestive pattern. Despite expressing \textit{more} non-neutral emotions (16.9\% vs.\ 5.7\%), AD patients showed reduced prosodic differentiation between emotion categories. Control speakers demonstrated substantial F0 modulation when expressing sadness ($\Delta$ = $-$3.45 semitones from neutral baseline), whereas AD speakers showed minimal F0 change ($\Delta$ = +0.11 semitones). This Group $\times$ Emotion interaction was statistically significant ($\beta$ = $-$5.20, $p$ = .023), with bootstrap validation (95\% CI: [$-$5.70, $-$0.22]). However, this finding should be interpreted with caution given the small sample sizes for non-neutral emotions, particularly sadness ($n$=5 utterances from 3 speakers in controls, $n$=15 from 11 speakers in AD). This preliminary pattern---which we tentatively term \textit{acoustic flattening}---may suggest that while AD patients perceive or intend to express emotions, their prosodic realization is attenuated. If replicated with larger samples, this finding would be consistent with prosodic impairments reported in AD \citep{themistocleous2020voice} and would have implications for emotion recognition systems, which may need to rely on multimodal cues rather than prosody alone when processing AD speech.

Our finding that expert annotators identified significantly more non-neutral emotions in AD speech, despite acoustic flattening, may help explain discrepancies with automated approaches. \citet{chou2025multimodal} found minimal emotion-related differences between AD and control groups using machine learning on acoustic features alone. While automated systems rely primarily on acoustic features that are attenuated in AD, human annotators integrate prosodic, linguistic, and pragmatic cues---such as lexical markers of empathy (``poor mother'') or face-saving laughter following word-finding difficulties. This suggests that emotion recognition in clinical speech may require multimodal approaches that go beyond acoustic analysis alone, and underscores the value of expert-annotated corpora for training such systems.

Voice quality measures (jitter, shimmer, HNR) did not significantly differentiate emotions after speaker normalization. This null finding may reflect disease-related baseline differences, or insufficient sample size for non-neutral emotions.

\subsection{Challenges in Emotion Annotation for Clinical Speech}

Our inter-rater reliability values (Fleiss' $\kappa$ = 0.23--0.31) are moderate by conventional standards but comparable to other spontaneous speech emotion corpora \citep{busso2008iemocap}. Several factors contribute to this difficulty: (1) emotion in spontaneous speech is inherently ambiguous, reflecting mixed emotions and mismatches between felt and expressed emotion; (2) atypical prosody in AD speech may be perceived as emotional even when patients intend neutral expression; (3) pragmatic functions of emotional expression differ in clinical populations, as documented in our calibration workshops (Section~\ref{sec:workshop}).

These challenges underscore the importance of releasing not only our annotations but also our annotation guidelines and calibration workshop materials.

\subsection{Null Results for Voice Quality Measures}

Voice quality measures (jitter, shimmer, HNR) did not significantly differentiate emotions after speaker normalization. Although HNR showed significant group differences at the speaker level (Table~\ref{tab:features}), this did not translate to emotion-level differentiation within speakers. 

We also examined H1-H2, which indexes phonation type (e.g., breathy vs.\ pressed voice (for more information, see \citet{keating2015acoustic}), but this measure is most reliably extracted from steady-state portions of vowels rather than from whole utterances. Our utterance-level extraction, which averages across consonants, pauses, and disfluencies, likely introduces substantial noise. Phone-level segmentation in v2.0 will enable more precise measurement of phonation-related features.

One possibility for the null results is that voice quality reflects stable, disease-related vocal changes rather than transient emotional states. AD-related neurodegeneration may affect laryngeal motor control \citep{themistocleous2020voice}, creating baseline voice quality differences that persist across emotional states. Alternatively, the sample size for non-neutral emotions may be insufficient to detect voice quality differences. Future work with larger non-neutral samples and phone-level acoustic analysis could clarify whether voice quality contributes to emotion differentiation in AD speech.

\section{Conclusion}
We presented DementiaBank-Emotion v1.0, the first multi-rater emotion annotation corpus for Alzheimer's disease speech. Our primary finding is that AD patients express significantly more non-neutral emotions than controls (16.9\% vs.\ 5.7\%; $p$ < .001). Exploratory acoustic analysis suggests a possible dissociation: control speakers showed substantial F0 modulation when expressing sadness ($\Delta$ = $-$3.45 semitones), whereas AD speakers showed minimal change ($\Delta$ = +0.11 semitones; interaction $p$ = .023). However, this \textit{acoustic flattening} pattern is based on limited samples (sadness: $n$=5 control, $n$=15 AD) and should be considered preliminary pending replication. Within AD speech, loudness differentiates emotion categories (higher for joy and surprise), indicating partially preserved emotion-prosody mappings.

We release the corpus, annotation guidelines, and calibration workshop materials to support future research. Planned extensions (v2.0) will incorporate longitudinal data and examination of relationships between emotional expression and cognitive severity. Additionally, ongoing phone-level realignment by trained phoneticians will enable fine-grained acoustic-phonetic analysis including phonation type, vowel space dynamics, and formant trajectories. This segmental analysis extends to filled pauses and discourse markers: AD patients use \textit{uh} more but \textit{um} less than controls \citep{yuan2020pauses}, and prosody disambiguates discourse marker functions \citep{jeong2017prosody}. Within the subjectification framework \citep{traugott1995subjectification,schiffrin1987discourse}, such markers index not only cognitive processing but also affective stance, connecting fine-grained phonetic analysis to the pragmatics of emotional expression when lexical resources are constrained.

\section*{Limitations}

Several limitations should be acknowledged.

\paragraph{Calibration Workshop for Control Data.} While both AD and control data come from the ADReSS 2020 Challenge dataset with matched demographics (54 speakers each, balanced for age and gender), AD data underwent three calibration workshops while control data did not due to time constraints. This led to higher ambiguity rates in AD (18.2\%) compared to control (1.2\%), as annotators for AD data developed shared implicit criteria through workshop discussions. Future versions should include calibration workshops for control annotation to ensure procedural consistency.

\paragraph{Class Imbalance.} The dataset is heavily skewed toward neutral ($>$83\% in AD, $>$94\% in control), with rare emotions (fear, disgust, anger) occurring in fewer than 10 utterances each. This class imbalance limits our ability to characterize the acoustic profiles of rare emotions and poses challenges for training supervised classifiers. As noted by a reviewer, pooling rare emotions into broader categories (e.g., ``negative valence'') may be necessary for statistical power in future acoustic analyses.

\paragraph{Task Context.} The Cookie Theft task represents a single communicative context---elicited picture description in an assessment setting. Emotional expression patterns may differ substantially in conversational, narrative, or real-world contexts. Our findings about face-saving laughter and preserved emotion-prosody mappings require replication across diverse communicative tasks before generalization.

\paragraph{Acoustic Analysis Granularity.} Acoustic features were extracted at the utterance level using original CHAT transcript timestamps, which is appropriate for global prosodic measures (F0, loudness) but precludes fine-grained analysis of within-utterance dynamics. Forced alignment was attempted but yielded approximately 20\% failure rates due to disfluencies characteristic of AD speech. Version 2.0 will incorporate manually corrected phone-level segmentation.

\paragraph{Perceived vs.\ Felt Emotion.} We did not have access to ground-truth emotional states. Our annotations reflect perceived emotion based on acoustic and linguistic cues, which may not correspond to speakers' actual felt emotions. This limitation is inherent to most emotion annotation work but is particularly salient in clinical populations where expressive and experiential components of emotion may dissociate.

\paragraph{Inter-Rater Reliability.} Our Fleiss' $\kappa$ values (0.23--0.31) are modest, though comparable to IEMOCAP ($\kappa$ = 0.27; \citealp{busso2008iemocap}). The moderate agreement reflects genuine ambiguity in clinical speech emotion rather than annotation noise, but users of this corpus should account for label uncertainty in downstream applications.

\section*{Ethical Considerations}

\paragraph{Data Source and Consent.} The speech data used in this study comes from the DementiaBank Pitt Corpus \citep{becker1994natural}, which was collected under IRB approval at the University of Pittsburgh. All participants provided informed consent at the time of original data collection. Our annotation work adds metadata to existing public data and does not involve new human subjects research; nevertheless, we obtained IRB exemption from the University of California, Irvine (IRB \#3795).

\paragraph{Privacy and Identifiability.} Although the DementiaBank corpus is publicly available for research purposes, speech recordings are inherently identifiable. We do not release any additional identifying information beyond what is already available in DementiaBank. Speaker IDs in our corpus correspond to the ADReSS Challenge identifiers, which are pseudonymized.

\paragraph{Vulnerable Population.} Individuals with Alzheimer's disease constitute a vulnerable population with diminished capacity for ongoing consent. We acknowledge the ethical complexity of annotating emotional states for individuals who cannot review or contest our labels. Our annotations reflect perceived emotion from external cues and should not be interpreted as claims about participants' internal states or used to make clinical judgments about individual patients.

\paragraph{Potential for Misuse.} Emotion recognition technology raises concerns about surveillance and manipulation. While our corpus is intended for research on clinical communication and assistive technology development, we acknowledge that emotion recognition systems could be misused. We encourage users to consider the ethical implications of downstream applications and to prioritize applications that benefit individuals with dementia and their caregivers.

\paragraph{Dual Use and AI Assistance.} We used Claude (Anthropic) for grammar checking and editing assistance in manuscript preparation. The scientific content, analysis, and interpretations are solely the responsibility of the human authors.

\section*{Data Availability}

The DementiaBank-Emotion corpus will be released through DementiaBank (\url{https://dementia.talkbank.org/}) upon publication.

\section*{Acknowledgments}
We appreciate the UC Noyce Initiative for their support. We thank the three anonymous reviewers for their constructive feedback, which substantially improved this paper. We used Claude (Anthropic) for grammar checking and editing assistance.

\bibliography{custom}

\newpage
\appendix
\section{Annotation Guidelines} 
\label{app:appendix_guideline}
The full annotation guidelines provided to the raters are included in the following pages. These guidelines detail the acoustic and linguistic criteria used to distinguish between Ekman's six basic emotions and the neutral state in clinical speech.


# Labeling guideline (ver 2.0, updated on June 9, 2024)

## 1. Introduction

- Goal
  - One task to complete for aim 3 is the annotation of utterances for emotion. This document discusses the method that should be used when annotating each.
- Task
  - Your task involves listening to conversations between a clinician (INV) and a patient (PAR), carefully reading the corresponding text, labeling it, and expressing your confidence level about the label. For procedure, see [4.2. Procedure](#).
  - We use 7 emotion categories (Ekman's 6 categories + neutral), which are defined in [5.1. Seven emotion categories](#).
- Data to label
  - A total of 495 utterances, including the header. For description, [see 4.1. Materials](#).
  - In the folder named your name
    - 1 text file in google sheet in the folder shared with you
    - 5 audio files in the same folder
- Timeline
  - Please do "first" first and then we can talk
- File location
  - First [first](#)
  - Second [second](#)
  - Third [third](#)

## 2. Key Terminology

### 2.1. Utterance

For purposes of this task, we define the term **utterance** as a single unit split by 10 milliseconds. In some cases, this will correspond to a single sentence without a pause while in others, this may actually be composed of more than one sentence. Occasionally, a single sentence is even split into two utterances.

### 2.2. Emotion

Emotion in this task refers to the discrete emotion shown by a speaker during an utterance. We use 7 categories (Ekman's 6 categories + neutral). The emotion is selected from the set of labels described below.

## 3. Data Description

For this task, we use an open dataset from *DementiaBank*, English Pitt Corpus (Becker et al., 1994). In the corpus, conversations between the clinician and patients are provided. Dementia patients are asked to describe the "cookie theft" picture below.

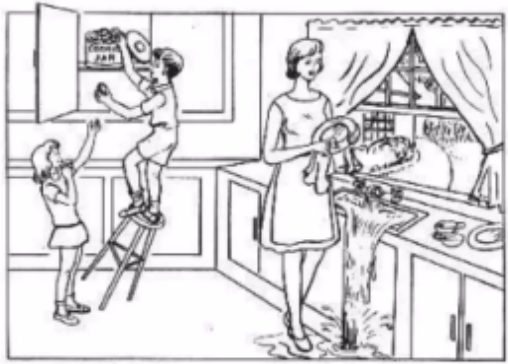

No

Some parts of an example conversation are as follows. As you can see, the clinician (INV) asks the patient (PAR) to describe the "cookie theft" picture, and the patient describes what they see.

| Speaker | Utterance |
|---|---|
| INV | what do you see happening in that picture ? |
| PAR | mhm . |
| INV | what's happening ? |
| PAR | well the kids is  robbin(g) a cookie jar . |
| PAR | and  the mother is  washin(g) dishes and forgettin(g) that |
| PAR | it's runnin(g) on the floor . |
| INV | mhm . |
| PAR | and his stool is about to dump him &=laughs . |

## 4. Annotation Procedure

### 4.1. Materials
- 1 text file: the_second_labeling.xlsx
- 7 audio files in the same folder
    - Each audio file, except for the first audio file (third_s108_122.wav) as it has lots of lines, includes specific lines and audios.
    - For example, in the file **third_s114_201.wav**, it covers conversations between S110 and S114, with utterances ranging from lines 123 to 201.

| Filename | Lines | Audios |
| --- | --- | --- |
| third_s108_122.wav | Lines 003-122 | S108 |
| third_s114_201.wav | Lines 123-201 | S110-114 |
| third_s125_306.wav | Lines 202-306 | S116-125 |
| third_s130_424.wav | Lines 307-424 | S126-132 |
| third_s139_504.wav | Lines 425-504 | S135-139 |
| third_s145_597.wav | Lines 505-686 | S140-145 |
| third_s156_686.wav | Lines 598-686 | S148-156 |

### 4.2. Procedure

1. For this task, you should have the spreadsheet open while listening to the audio. You should select a quiet place to work and use headphones to ensure that you can clearly hear the audio.

2. After listening to each utterance, pause the recording, <u>then enter the emotion label into the Emotion Class column in the spreadsheet under column D.</u> Additionally, indicate <u>your confidence level about the label in column E ("How sure are you?")</u>. You can then resume the recording to examine the next utterance.
3. If unclear, listen to previous and/or following utterance(s).
4. <u>Each utterance should be given a single label</u>. This label may be based on the words that the participant produces, the way in which they speak, or both.

## 5. Emotion labels

### 5.1. Seven Emotion categories

For labeling, we are using a set composed of Ekman's 6 universal emotions + a **neutral** label. This label set is:

1. **anger**: the speaker is angry, upset, and reveals this through words, tone or both.
2. **disgust**: the speaker is disgusted, unpleasant, offensive or revolted.
3. **fear**: the speaker is afraid of something.
4. **neutral:** (no clear emotion)--the speaker does neither demonstrate any emotions nor any opinions. They may be stating, describing or providing information about what they are seeing in the picture.
5. **joy**: the speaker is happy, having a good time, or otherwise enjoying something.
6. **sadness**: It conveys loss, helplessness, or disappointment.
7. **surprise**: something surprising has happened, the speaker is suddenly given new unexpected information.

### 5.2. How to decide which emotion label to select

Choosing the right label for an utterance is usually straightforward, but there are times when it can be a bit tricky. In general, follow these guidelines:

- If an utterance contains no obvious emotional information, give it a label of **neutral**. Otherwise, put specific labels below.
- If an utterance is mostly neutral but includes a segment with clear emotional content, label it based on the emotion expressed in that non-neutral segment.
- If an utterance contains two emotions, do the following:
    - If one emotion seems much stronger than the other, choose the stronger emotion
    - If one emotion dominates the utterance, choose the dominant emotion

**Caveat!!!**

- Exercise caution regarding personal impressions and the use of specific words. For example, avoid labeling based solely on words like "bitchy," as these may not accurately reflect the patient's emotions. Connotations of such words and cultural factors can influence the labeling.
- Understanding the overall speech style of patients can be beneficial. Remember that the delivery of speech matters more than the specific words used. Focus on the emotions expressed by patients rather than speculating on the causes of these emotions.

**Specific labels with speech features:**
1. **Joy**
    a. "Laughing" is a strong indicator, but carefully put it as it can overlap sarcasm.
    b. When patients feel joy, their speech features high volume, moderate volume variance, and/or high pitch variance.
    c. Their speech rate is moderate, with long durations for both their speech and pauses.
    d. When clinicians encourage patients to speak, it's common for their speech to become exaggerated, with a higher or rising pitch and intensity. The speech style often takes on a humorous tone.
2. **Sadness**
    a. If patients muddle and/or speak unclearly, vaguely, use "sadness," but carefully label as it can overlap fear.
    b. When patients feel sadness, their speech features low volume, high volume variance, and/or low pitch variance. Their overall pitch is low.
    c. Their speech rate is slow, with long durations for both their speech and pauses. Many pauses are found.
3. **Fear**
    a. When patients express fear about something, use "fear." Fear triggers include loss of visibility of surroundings, social interaction and/or rejection, death and dying, etc.
    b. When patients feel fear, their speech features high volume, low volume variance, and/or high pitch variance. Their overall pitch is high.
    c. Their speech rate is fast, with short durations for both their speech and pauses. Not many pauses.
4. **Disgust**
    a. When patients express contempt about something, use "disgust," but carefully label as it can overlap anger.
5. **Anger**
    a. When patients feel anger, their speech features high volume, high volume variance, and/or low pitch variance. Their overall pitch is low.
    b. Their speech rate is fast, with short durations for both their speech and pauses. Few pauses are found.

6. **Surprise**
      a. When something unexpected occurs, patients' pitch changes dramatically, but carefully label as it can overlap joy.
      b. Gasping can also serve as an indicator.
   7. **Neutral**
      a. When there is no specific feature observed, put "neutral."
      b. When patients explain and/or describe something step by step, patients' speech tends to be flat, with a monotonous tone.
      c. Rather than words that deliver emotions, descriptive words are used without specific opinions.



\onecolumn
\section{Full Ambiguous Utterances}
\label{app:ambiguous_list}
\input{appendix_ambiguous_full}

\newpage
\section{Acoustic Features by Emotion}
\label{app:acoustic_emotion}

\begin{figure}[h]
\centering
\includegraphics[width=\columnwidth]{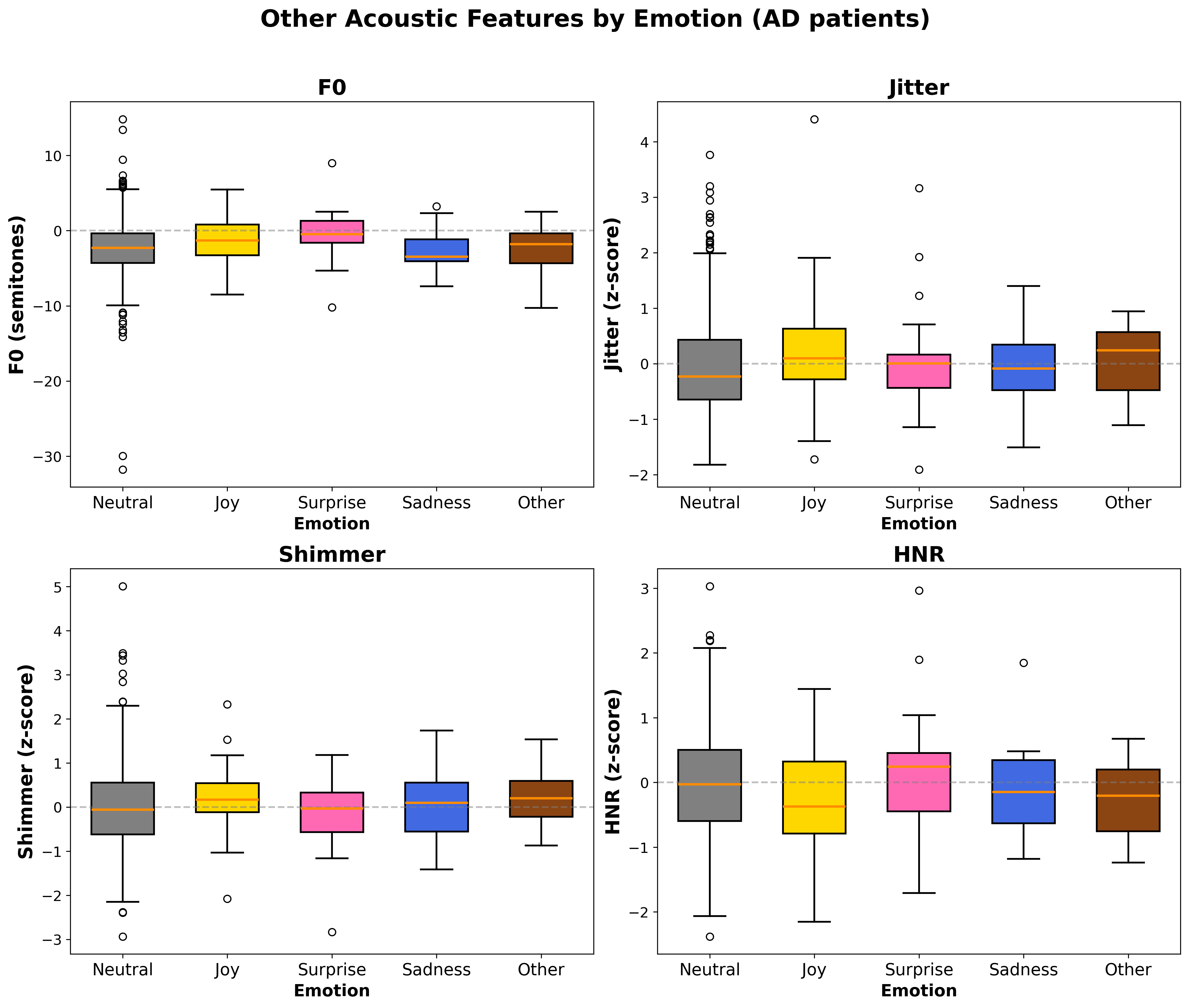}
\caption{Speaker-normalized acoustic features by emotion category (AD patients only). Error bars indicate 95\% confidence intervals.}
\label{fig:appendix_acoustic}
\end{figure}

\end{document}

%% file: appendix_ambiguous_full.tex
\begin{longtable}{@{}l p{5.5cm} rr rr rr@{}}
\caption{All ambiguous utterances ($n$ = 146) from the ADReSS 2020 Challenge dataset with acoustic features, sorted by F0 variance $z$-score (descending). Speaker IDs (S0xx) correspond to ADReSS identifiers. eGe = eGeMAPS (F0 in st from 27.5 Hz; F0 var = stddevNorm; Loudness in sone). spk/$z$ = speaker-normalized (F0 in st from speaker mean; others $z$-scored within speaker).} \label{tab:full_ambiguous} \\
\toprule
& & \multicolumn{2}{c}{\textbf{F0 (st)}} & \multicolumn{2}{c}{\textbf{F0 var}} & \multicolumn{2}{c}{\textbf{Loudness}} \\
\cmidrule(lr){3-4} \cmidrule(lr){5-6} \cmidrule(lr){7-8}
\textbf{Spk} & \textbf{Utterance} & eGe & spk & eGe & $z$ & eGe & $z$ \\
\midrule
\endfirsthead

\multicolumn{8}{c}{\textit{(continued from previous page)}} \\
\toprule
& & \multicolumn{2}{c}{\textbf{F0 (st)}} & \multicolumn{2}{c}{\textbf{F0 var}} & \multicolumn{2}{c}{\textbf{Loudness}} \\
\cmidrule(lr){3-4} \cmidrule(lr){5-6} \cmidrule(lr){7-8}
\textbf{Spk} & \textbf{Utterance} & eGe & spk & eGe & $z$ & eGe & $z$ \\
\midrule
\endhead

\midrule
\multicolumn{8}{r}{\textit{(continued on next page)}} \\
\endfoot

\bottomrule
\endlastfoot

S095 & ``xxx anything else ?'' & 23.6 & +3.8 & 0.52 & +2.79 & 0.09 & +0.74 \\
S093 & ``and (.) for some reason she must have been upset abo...'' & 34.0 & +3.7 & 0.52 & +2.33 & 0.03 & -0.62 \\
S126 & ``looks +/.'' & 29.7 & +2.3 & 0.34 & +1.83 & 0.13 & +1.74 \\
S126 & ``(.) and there's somethin(g) else over there .'' & 20.5 & -6.9 & 0.33 & +1.67 & 0.03 & -1.05 \\
S124 & ``and the sink's runnin(g) xxx .'' & 21.4 & -1.7 & 0.31 & +1.47 & 0.03 & -0.83 \\
S111 & ``his k[er]@u .'' & 33.9 & -0.5 & 0.27 & +1.41 & 0.15 & -1.11 \\
S086 & ``he's tryin(g) to kill himself xxx .'' & 32.3 & +7.1 & 0.32 & +1.35 & 0.13 & +2.37 \\
S093 & ``but she would have gotten hold of him and saved him ...'' & 30.6 & +0.3 & 0.38 & +1.34 & 0.06 & +0.31 \\
S082 & ``it splashed from the sink but not from (.) from +...'' & 27.9 & +0.9 & 0.39 & +1.32 & 0.07 & -0.35 \\
S072 & ``(..) she is not paying any attention to her kids .'' & 24.6 & +1.9 & 0.42 & +1.30 & 0.53 & -1.62 \\
S082 & ``and all of a sudden somebody turned over a dish .'' & 30.5 & +3.5 & 0.38 & +1.29 & 0.09 & -0.28 \\
S090 & ``well let's see .'' & 32.9 & +3.8 & 0.46 & +1.28 & 0.06 & -0.98 \\
S107 & ``(..) xxx .'' & 28.6 & +0.6 & 0.13 & +1.14 & 0.08 & -0.43 \\
S125 & ``oh .'' & 29.6 & -4.0 & 0.15 & +1.14 & 0.14 & +2.20 \\
S080 & ``+< okay .'' & 34.6 & +17.3 & 0.31 & +1.11 & 0.37 & +1.45 \\
S089 & ``so she will find her .'' & 26.7 & -2.4 & 0.29 & +1.10 & 0.14 & +1.10 \\
S090 & ``it looks like +...'' & 35.4 & +6.2 & 0.44 & +1.08 & 0.07 & -0.92 \\
S095 & ``and then she's not even lookin(g) at them .'' & 16.0 & -3.8 & 0.27 & +1.05 & 0.02 & -0.52 \\
S107 & ``xxx .'' & 28.5 & +0.5 & 0.12 & +1.00 & 0.29 & +0.91 \\
S082 & ``and the mother does not see it because she's inside ...'' & 27.3 & +0.3 & 0.35 & +0.96 & 0.09 & -0.28 \\
S126 & ``(.) there's another girl (.) look like .'' & 25.9 & -1.5 & 0.29 & +0.93 & 0.04 & -0.61 \\
S126 & ``(.) some little knots or somethin(g) .'' & 25.5 & -1.9 & 0.29 & +0.83 & 0.04 & -0.57 \\
S111 & ``her .'' & 33.6 & -0.8 & 0.23 & +0.82 & 1.05 & +1.05 \\
S093 & ``maybe dropped her dish \&=laughs .'' & 36.3 & +6.0 & 0.31 & +0.81 & 0.08 & +1.04 \\
S139 & ``(..) xxx .'' & 26.8 & -1.6 & 0.29 & +0.80 & 0.02 & -1.23 \\
S125 & ``(..) that's about it .'' & 35.6 & +2.0 & 0.14 & +0.80 & 0.02 & -0.60 \\
S003 & ``and he's getting a cookie and he's sharing a cookie ...'' & 31.9 & -1.3 & 0.21 & +0.76 & 0.55 & -0.41 \\
S126 & ``I see a girl standin(g) there or somethin(g) or other .'' & 27.3 & -0.1 & 0.28 & +0.74 & 0.05 & -0.36 \\
S111 & ``+< xxx ample xxx rice \&=clears:throat discharged .'' & 34.8 & +0.4 & 0.23 & +0.73 & 0.72 & +0.27 \\
S126 & ``(loo)ks like somebody took some pencils or somethin(...'' & 25.8 & -1.7 & 0.28 & +0.73 & 0.08 & +0.44 \\
S125 & ``and the mother is spilling the water .'' & 34.3 & +0.7 & 0.13 & +0.72 & 0.02 & -0.46 \\
S093 & ``I can't believe she's upset about the kids (be)cause...'' & 24.9 & -5.3 & 0.29 & +0.71 & 0.03 & -0.72 \\
S093 & ``in other words it's +...'' & 25.8 & -4.5 & 0.29 & +0.71 & 0.03 & -0.80 \\
S124 & ``\&=sighs window's open .'' & 22.4 & -0.7 & 0.24 & +0.71 & 0.05 & -0.30 \\
S082 & ``I'm too too trying to get too much out\_of it .'' & 25.5 & -1.5 & 0.33 & +0.70 & 0.04 & -0.50 \\
S082 & ``except that it did did not dry it up .'' & 26.7 & -0.3 & 0.33 & +0.70 & 0.05 & -0.44 \\
S108 & ``mhm .'' & 35.6 & +3.7 & 0.26 & +0.67 & 0.16 & -1.11 \\
S082 & ``+ (.) but not getting anything that you'll want want...'' & 30.7 & +3.7 & 0.32 & +0.65 & 0.13 & -0.08 \\
S090 & ``oh +...'' & 24.9 & -4.2 & 0.38 & +0.58 & 0.17 & -0.30 \\
S079 & ``and she's she has has +/.'' & 24.8 & +0.0 & 0.32 & +0.57 & 0.37 & -0.50 \\
S082 & ``+, a weak image, so to speak .'' & 33.9 & +6.9 & 0.31 & +0.55 & 0.29 & +0.63 \\
S082 & ``+< but an etch you would say it in a little .'' & 27.4 & +0.4 & 0.31 & +0.52 & 0.15 & +0.02 \\
S095 & ``and his stool is fallin(g) over .'' & 38.3 & +18.4 & 0.19 & +0.51 & 0.04 & -0.10 \\
S082 & ``+< and one one of the kids is gonna get a crack on t...'' & 32.7 & +5.7 & 0.30 & +0.44 & 0.15 & +0.02 \\
S082 & ``+< \&=laughs xxx !'' & 35.2 & +8.2 & 0.30 & +0.43 & 1.06 & +4.13 \\
S111 & ``this is oh I can just +...'' & 32.2 & -2.2 & 0.20 & +0.43 & 0.43 & -0.42 \\
S125 & ``there's a cookie jar .'' & 34.6 & +1.0 & 0.12 & +0.43 & 0.03 & -0.44 \\
S093 & ``she she's deciding that if she did see them she's de...'' & 34.9 & +4.6 & 0.25 & +0.41 & 0.05 & +0.08 \\
S126 & ``some of xxx and things .'' & 28.3 & +0.9 & 0.26 & +0.40 & 0.03 & -0.81 \\
S090 & ``(..) oh I can't read +...'' & 28.1 & -1.0 & 0.35 & +0.37 & 0.05 & -1.06 \\
S096 & ``no I can't get this very well, clear .'' & 28.1 & -3.9 & 0.26 & +0.36 & 0.16 & -0.89 \\
S089 & ``and xxx the mother washes dryin(g) the dishes .'' & 30.0 & +0.8 & 0.24 & +0.32 & 0.06 & -0.50 \\
S020 & ``that'd be pretty difficult to look out the window an...'' & 23.9 & +0.9 & 0.22 & +0.29 & 0.11 & -1.30 \\
S124 & ``xxx the mother over there she's doing the dishes .'' & 21.6 & -1.4 & 0.20 & +0.26 & 0.02 & -0.92 \\
S111 & ``xxx .'' & 39.1 & +4.7 & 0.19 & +0.25 & 0.66 & +0.12 \\
S051 & ``and what else ?'' & 24.6 & -0.6 & 0.33 & +0.25 & 0.57 & +0.36 \\
S017 & ``yeah that's it .'' & 37.0 & +4.9 & 0.24 & +0.23 & 0.42 & +1.00 \\
S107 & ``boing@o .'' & 27.1 & -0.9 & 0.08 & +0.21 & 0.09 & -0.34 \\
S090 & ``+< xxx is a +/.'' & 28.3 & -0.8 & 0.34 & +0.21 & 0.43 & +1.49 \\
S107 & ``mhm .'' & 39.7 & +11.6 & 0.08 & +0.20 & 0.11 & -0.23 \\
S111 & ``+< xxx .'' & 35.5 & +1.1 & 0.19 & +0.19 & 0.84 & +0.56 \\
S084 & ``you\_know it I I excuse me but you\_know I I was +...'' & 30.1 & +2.9 & 0.30 & +0.19 & 0.16 & -0.12 \\
S095 & ``and the little girl is beggin(g) him to give her one .'' & 36.7 & +16.8 & 0.14 & +0.13 & 0.04 & -0.20 \\
S107 & ``xxx .'' & 27.3 & -0.8 & 0.07 & +0.12 & 0.04 & -0.64 \\
S095 & ``+< and I will tell you what's +/.'' & 40.5 & +20.6 & 0.13 & +0.12 & 0.22 & +3.38 \\
S126 & ``look to me like the same except them things up there .'' & 28.1 & +0.7 & 0.24 & +0.09 & 0.05 & -0.45 \\
S124 & ``maybe the wind is blowing in .'' & 24.2 & +1.2 & 0.18 & +0.09 & 0.11 & +1.19 \\
S093 & ``and he's handing a cookie down to her .'' & 30.2 & -0.0 & 0.21 & +0.08 & 0.03 & -0.83 \\
S126 & ``(.) xxx .'' & 26.2 & -1.3 & 0.24 & +0.03 & 0.05 & -0.34 \\
S126 & ``(.) some kind of a xxx pan or somethin(g) .'' & 23.9 & -3.5 & 0.24 & +0.02 & 0.06 & -0.24 \\
S093 & ``and he might just pull all the cookie jar with him w...'' & 40.3 & +10.0 & 0.19 & -0.03 & 0.13 & +2.59 \\
S082 & ``it's so +...'' & 35.1 & +8.1 & 0.25 & -0.10 & 0.07 & -0.33 \\
S082 & ``sometimes I I see it very clear and and other times ...'' & 27.8 & +0.8 & 0.25 & -0.11 & 0.08 & -0.29 \\
S108 & ``yes .'' & 35.6 & +3.7 & 0.18 & -0.13 & 0.47 & -0.37 \\
S082 & ``and I guess in the the picture here that the mother ...'' & 24.4 & -2.6 & 0.25 & -0.15 & 0.07 & -0.36 \\
S126 & ``(.) and that girl is there .'' & 25.8 & -1.6 & 0.23 & -0.18 & 0.06 & -0.12 \\
S082 & ``\&=clears:throat well \&=clears:throat the kids are ...'' & 27.8 & +0.8 & 0.25 & -0.19 & 0.04 & -0.50 \\
S093 & ``but I don't think she would have let them go ahead i...'' & 27.9 & -2.4 & 0.17 & -0.19 & 0.04 & -0.50 \\
S086 & ``that's it ?'' & 30.5 & +5.3 & 0.13 & -0.28 & 0.08 & +0.73 \\
S095 & ``two cups and and a dish finished .'' & 41.9 & +22.1 & 0.08 & -0.29 & 0.04 & -0.16 \\
S095 & ``their mama is doin(g) the dishes .'' & 37.3 & +17.5 & 0.08 & -0.29 & 0.05 & +0.08 \\
S082 & ``they they are going to get some cookies from the coo...'' & 24.3 & -2.7 & 0.23 & -0.34 & 0.09 & -0.25 \\
S029 & ``she is apparently so distract daydreaming that she c...'' & 28.6 & -0.0 & 0.22 & -0.34 & 0.16 & -1.08 \\
S093 & ``+, threesome but and the kitchen would be a mess .'' & 27.3 & -3.0 & 0.14 & -0.37 & 0.04 & -0.31 \\
S096 & ``I see a tad bit .'' & 33.4 & +1.4 & 0.20 & -0.38 & 0.39 & +1.06 \\
S124 & ``hmhunh .'' & 19.6 & -3.4 & 0.14 & -0.45 & 0.02 & -1.10 \\
S093 & ``she's washing the dishes .'' & 41.8 & +11.5 & 0.13 & -0.45 & 0.05 & +0.11 \\
S107 & ``(...) towel .'' & 27.4 & -0.6 & 0.04 & -0.46 & 0.02 & -0.79 \\
S090 & ``\&=sings .'' & 25.5 & -3.6 & 0.26 & -0.46 & 0.14 & -0.49 \\
S093 & ``well the children are climbing up and he's about to ...'' & 36.2 & +6.0 & 0.13 & -0.48 & 0.05 & +0.13 \\
S118 & ``+< oh (.) I don't know .'' & 31.7 & +1.4 & 0.18 & -0.50 & 0.74 & +1.53 \\
S107 & ``windows windows +...'' & 28.4 & +0.4 & 0.04 & -0.50 & 0.11 & -0.19 \\
S124 & ``stool's on that step xxx .'' & 17.3 & -5.7 & 0.13 & -0.52 & 0.03 & -0.70 \\
S126 & ``(..) I see this they all look looked about the same ...'' & 27.6 & +0.1 & 0.21 & -0.53 & 0.05 & -0.45 \\
S095 & ``the water's runnin(g) over the sink .'' & 13.4 & -6.4 & 0.04 & -0.56 & 0.02 & -0.53 \\
S093 & ``maybe she did turn and look at them and \&=laughs th...'' & 31.4 & +1.1 & 0.12 & -0.58 & 0.06 & +0.25 \\
S051 & ``and in looking out the window why she's lettin(g) he...'' & 26.9 & +1.7 & 0.23 & -0.58 & 0.83 & +1.57 \\
S084 & ``did it +/?'' & 34.2 & +7.0 & 0.22 & -0.62 & 0.38 & +0.85 \\
S107 & ``girl assisting boy with cookie jar .'' & 27.3 & -0.8 & 0.03 & -0.63 & 0.07 & -0.47 \\
S126 & ``(.) I don't see xxx .'' & 30.7 & +3.2 & 0.20 & -0.65 & 0.07 & +0.17 \\
S082 & ``and the kids then just +...'' & 24.4 & -2.6 & 0.20 & -0.65 & 0.05 & -0.42 \\
S107 & ``(.) garage .'' & 27.5 & -0.5 & 0.03 & -0.70 & 0.03 & -0.74 \\
S126 & ``I don't know .'' & 28.9 & +1.5 & 0.19 & -0.77 & 0.10 & +0.89 \\
S125 & ``and the boy is toppling off (.) a stool .'' & 33.3 & -0.3 & 0.06 & -0.78 & 0.05 & +0.09 \\
S107 & ``xxx .'' & 28.0 & -0.0 & 0.02 & -0.81 & 0.05 & -0.62 \\
S107 & ``dryin(g) dishes .'' & 26.5 & -1.5 & 0.02 & -0.81 & 0.05 & -0.63 \\
S095 & ``she wants to eat it .'' & 0.0 & -19.9 & 0.00 & -0.82 & 0.02 & -0.51 \\
S095 & ``and she's pointin(g) to her mouth .'' & 0.0 & -19.9 & 0.00 & -0.82 & 0.02 & -0.60 \\
S095 & ``it's a nice yard out there .'' & 0.0 & -19.9 & 0.00 & -0.82 & 0.02 & -0.58 \\
S095 & ``that's a mess .'' & 0.0 & -19.9 & 0.00 & -0.82 & 0.02 & -0.60 \\
S095 & ``dryin(g) dishes .'' & 0.0 & -19.9 & 0.00 & -0.82 & 0.02 & -0.58 \\
S095 & ``I think she's lookin(g) out the window .'' & 0.0 & -19.9 & 0.00 & -0.82 & 0.02 & -0.60 \\
S107 & ``chair .'' & 28.4 & +0.4 & 0.02 & -0.83 & 0.04 & -0.69 \\
S080 & ``can't see anything else .'' & 0.0 & -17.3 & 0.00 & -0.90 & 0.02 & -1.00 \\
S080 & ``no .'' & 0.0 & -17.3 & 0.00 & -0.90 & 0.02 & -0.94 \\
S080 & ``no .'' & 0.0 & -17.3 & 0.00 & -0.90 & 0.02 & -0.98 \\
S086 & ``there's two cups and a saucer or a plate maybe .'' & 25.1 & -0.1 & 0.06 & -0.91 & 0.05 & -0.14 \\
S107 & ``xxx .'' & 27.9 & -0.1 & 0.02 & -0.91 & 0.18 & +0.21 \\
S100 & ``I don't know \&=laughs .'' & 32.6 & +2.9 & 0.04 & -0.92 & 0.03 & -1.09 \\
S124 & ``(.) and the little girl is reachin(g) up for a cookie .'' & 24.0 & +0.9 & 0.09 & -0.93 & 0.04 & -0.55 \\
S125 & ``mhm .'' & 37.3 & +3.7 & 0.05 & -0.96 & 0.04 & -0.18 \\
S081 & ``the girl's laughin(g) at her brother because he went...'' & 21.5 & -0.9 & 0.10 & -0.98 & 0.42 & +1.10 \\
S082 & ``and maybe he has +...'' & 28.3 & +1.3 & 0.17 & -1.06 & 0.04 & -0.49 \\
S079 & ``like the the mother is near the girl .'' & 23.3 & -1.4 & 0.16 & -1.09 & 0.45 & -0.02 \\
S039 & ``I do I start ?'' & 27.9 & -2.1 & 0.13 & -1.13 & 0.50 & -0.57 \\
S086 & ``there's cookies in the jar up in the pantry I suppose .'' & 25.4 & +0.2 & 0.03 & -1.17 & 0.02 & -0.99 \\
S126 & ``look like a little kid the same .'' & 27.4 & -0.0 & 0.17 & -1.17 & 0.05 & -0.29 \\
S090 & ``+< did you ?'' & 30.3 & +1.1 & 0.17 & -1.20 & 0.77 & +3.73 \\
S135 & ``what is she +//?'' & 24.4 & +0.0 & 0.03 & -1.23 & 0.02 & -0.86 \\
S126 & ``look like some a little girl is in there .'' & 29.6 & +2.1 & 0.16 & -1.25 & 0.17 & +2.83 \\
S124 & ``the boy's getting the cookies .'' & 22.4 & -0.7 & 0.06 & -1.26 & 0.04 & -0.66 \\
S100 & ``oh (.) +...'' & 17.3 & -12.3 & 0.01 & -1.26 & 0.02 & -1.24 \\
S076 & ``mhm a\_lot\_of things are happening .'' & 29.9 & +1.2 & 0.23 & -1.31 & 1.43 & +0.20 \\
S081 & ``she's looking out the window .'' & 23.0 & +0.7 & 0.05 & -1.31 & 0.14 & -1.36 \\
S125 & ``and the what else ?'' & 30.5 & -3.1 & 0.03 & -1.35 & 0.02 & -0.60 \\
S093 & ``and he's getting cookies .'' & 30.5 & +0.2 & 0.00 & -1.39 & 0.03 & -0.71 \\
S093 & ``and she's telling him to sh@o be quiet so mother won...'' & 31.2 & +0.9 & 0.00 & -1.42 & 0.03 & -0.87 \\
S093 & ``I think she would have turned .'' & 0.0 & -30.3 & 0.00 & -1.42 & 0.02 & -0.96 \\
S126 & ``I don't see nothin(g) else .'' & 33.2 & +5.8 & 0.15 & -1.43 & 0.02 & -1.22 \\
S139 & ``mhm .'' & 33.1 & +4.7 & 0.00 & -1.52 & 0.02 & -1.10 \\
S139 & ``he has a cookie in his hand .'' & 0.0 & -28.4 & 0.00 & -1.52 & 0.02 & -1.11 \\
S135 & ``she +...'' & 0.0 & -24.4 & 0.00 & -1.59 & 0.01 & -1.02 \\
S089 & ``mhm .'' & 35.9 & +6.7 & 0.15 & -1.61 & 0.11 & +0.45 \\
S082 & ``and all over the floor .'' & 23.0 & -4.0 & 0.08 & -2.01 & 0.08 & -0.33 \\
S090 & ``mhm .'' & 23.5 & -5.7 & 0.07 & -2.02 & 0.06 & -1.03 \\
S082 & ``man !'' & 0.0 & -27.0 & 0.00 & -2.85 & 0.02 & -0.59 \\
\end{longtable}